\documentclass[11pt,a4paper]{article}
\usepackage[hyperref]{acl2017}
\usepackage{times}
\usepackage{latexsym}
\usepackage{graphicx}
\usepackage{url}

\aclfinalcopy % Uncomment this line for the final submission
\title{Predicting the Law Area and Decisions of French Supreme Court Cases}

\author{Octavia-Maria \c{S}ulea\textsuperscript{1}, Marcos Zampieri\textsuperscript{2}, Mihaela Vela\textsuperscript{3}, Josef van Genabith\textsuperscript{3,4} \\
  \textsuperscript{1}University of Bucharest, Romania \\
  \textsuperscript{2}University of Wolverhampton, United Kingdom \\
  \textsuperscript{3}Saarland University, Germany \\
  \textsuperscript{4}German Research Center for Artificial Intelligence (DFKI), Germany \\
  {\tt mary.octavia@gmail.com, marcos.zampieri@uni-koeln.de} \\
  {\tt m.vela@mx.uni-saarland.de, josef.vangenabith@uni-saarland.de} 
\\
  }

\date{}

\begin{document}
\maketitle
\begin{abstract}
In this paper, we investigate the application of text classification methods to predict the law area and the decision of cases judged by the French Supreme Court. We also investigate the influence of the time period in which a ruling was made over the textual form of the case description and the extent to which it is necessary to mask the judge's motivation for a ruling to emulate a real-world test scenario. We report results of 96\% f1 score in predicting a case ruling, 90\% f1 score in predicting the law area of a case, and 75.9\% f1 score in estimating the time span when a ruling has been issued using a linear Support Vector Machine (SVM) classifier trained on lexical features.
\end{abstract}

\section{Introduction}

Text classification methods have been used in a wide range of NLP tasks. This includes predicting information about authors of texts, such as age \cite{nguyen2013}, gender \cite{ciobanu2017including}, personality traits \cite{sulea2}, and native language \cite{gebre2013}, estimating the period in which a text was published \cite{niculae2014temporal}, the amount of subjectivity or sentiment expressed in texts \cite{balahur14}, and detecting pastiche \cite{dinu2012pastiche}, plagiarism \cite{barron2013plagiarism}, and influences from other authors \cite{ganascia14}. Classic machine learning algorithms such as Multinomial Naive Bayes and SVMs proved to be very reliable for these tasks, achieving high performance.

In this paper, we apply text classification methods to legal documents. We explore the use of bag of words (BOW) and linear SVM classifiers in predicting a case's ruling, law area, and the date in which a ruling was issued. We apply these methods to a large corpus of court rulings issued by the French Supreme Court with over 126,000 documents, spanning from the 1800s until the present day. 

To the best of our knowledge, several NLP tasks have been carried out on legal texts, most notably text summarization \cite{farzindar2004legal,galgani12}, however, as evidenced in Section \ref{sec:related}, the use of text classification to predict court rulings is an under-explored area. The recent study by \newcite{aletras2016predicting} on predicting decisions of the European Court of Human Rights (ECHR) is among the few examples of such attempts. 

\section{Related Work}
\label{sec:related}

In the legal domain, text classification has been more important to forensics \cite{de2001mining,dark-triad,rada15} than to predict information in legal texts such as case descriptions, rulings, and court decisions. General NLP methods, on the other hand, have played an important role in the intersection between artificial intelligence and law, a vibrant sub-area of research with international associations (e.g. IAAIL\footnote{\url{http://www.iaail.org/}}) and a number of specialized scientific conferences and workshops.

\newcite{Palau:2009} investigate the extent to which one can automatically identify argumentative propositions in legal text, along with their argumentative function and structure. They use a corpus containing legal texts extracted from the European Court of Human Rights (ECHR) and classify argumentative vs. non-argumentative sentences with an accuracy of 80\%. 

\newcite{Boella:2011} present a classification approach to identify the relevant domain to which a specific legal text belongs. Using TF-IDF weighting and Information Gain for feature selection and SVM for classification, reporting an f1-measure of 76\% for the identification of the domains related to a legal text and 97.5\% for the correct classification of a text into a specific domain.
 
The studies by \newcite{farzindar2004legal} and by \newcite{galgani12} apply computational methods for the automatic summarization of legal texts. Such applications are developed to help law professionals in speeding up their work by providing shorter summaries of very long documents which are abundant in legal processes. 

Studies applying text classification to legal documents include \newcite{hachey2006extractive}, which proposed a system of classifying sentences for automatic court rulings summarization, and  \newcite{gonccalves2005evaluating}, which used BOW, POS tags, and TF-IDF to classify legal text in 3,000 categories, based on a taxonomy of legal concepts. Authors of this study reported performance of 64\% and 79\% f1 score. 

A few papers have been published on court ruling prediction. This includes the work by \newcite{Katz14}, using extremely randomized trees, reporting 70\% accuracy in predicting the US Supreme Court's behavior and, more recently, \newcite{aletras2016predicting} proposed a computational method to predict decisions of the ECHR and reported 78\% accuracy as their highest score.

To the best of our knowledge, so far most work on predicting court rulings has been carried out on English data. No work has yet been carried out on French, such as the Supreme Court decisions we analyze in this paper. Moreover, to the best of our knowledge, previous work on court rule prediction did not take a temporal dimension into account and our work fills this gap. 

Finally, another innovative aspect of our work is the masking step described in \ref{sec:masking}. French High Court rulings contain (near) explicit mentions of our targeted predictions in the running text of the ruling (e.g. law area, court ruling, and time issued). In order to simulate a realistic application scenario where a text classification-based system is supposed to make a prediction on ``draft'' case description data that do not mention the predicted variables, we present a method of automatically masking such data based on feature ranking on the full data. All our experiments reported in this paper are carried out with masked data sets. 

\section{Data}
\label{sec:methods}

%\subsection{Corpus}

We use a diachronic collection of rulings from the French supreme court ({\em Court de Cassation}).\footnote{\url{https://www.courdecassation.fr/about_the_court_9256.html}} The complete collection\footnote{\url{https://www.legifrance.gouv.fr}} contains 131,830 documents, each consisting of a unique ruling and metadata formatted in XML. Common metadata available in most documents includes: law area, time stamp, case ruling (e.g. \emph{cassation},  \emph{rejet}, \emph{non-lieu}, etc.), case description, and cited laws. %In the following, we denote each XML document by the term court ruling (CR). Each CR contains 
In our supervised learning approach we use the metadata provided as `natural' labels to be predicted by the machine learning system. In order to simulate realistic test scenarios, we identify and mask all mentions from the training and test data that refer to our target prediction classes. In a pre-processing step we remove all surface forms of the words within the labels from the text data used to derive the predictive features. 

All duplicate and incomplete entries in the dataset were excluded resulting in a corpus comprising 126,865 unique court rulings, each containing a case description and four different types of labels: a law area, the date of ruling, the case ruling itself, and a list of articles and laws cited within the description. %These entries were stored in .csv format, an entry example of which can be seen below.
%We investigate the extent to which the ruling and law area can be predicted from the description. And how much of the description needed to be \emph{masked} to be closer to a legal "draft" scenario.

%For each of the two tasks investigated in the paper (law area and ruling prediction) we separately identify and remove all mentions in the case description that directly or indirectly give away the target class, using word unigram feature selection.

\subsection{Tasks and Labels}
\label{sec:label}

In this section we present the process of defining labels in the dataset for the three tasks presented in this paper. The tasks and the respective section of the paper containing the results are summarized as follows:

\begin{enumerate}
\item Predicting the law area of a case (Section \ref{sec:area}).
\item Predicting the court ruling based on the respective case description (Section \ref{sec:ruling}).
\item Estimating when a case description and a ruling were issued (Section \ref{sec:temporal}).
\end{enumerate}

%Many classifiers typically used in NLP tend to favor the majority class \cite{weiss,vanhulse} and often work with features extracted over the texts' vocabulary. We therefore worked to obtain fewer and more balanced labels as well as a smaller vocabulary. 

%Since it is known (\cite{weiss},\cite{vanhulse}) that binary classifiers have a bias toward the majority class and since \cite{wei} indicates that balanced training sets lead to better (test) performance even when the test sets are unbalanced, 

\noindent To reduce the feature and label space, we first removed accents and punctuation and lowercased all words in the description and ruling. Further pre-processing was needed to reduce the label space for each task. 
%Specifically, for the law area task, out of 17 initial unique labels, one (\emph{avis}) was the same as the ruling label, and appeared in only 34 entries, and was probably an annotation error, and 8 labels appeared in the corpus less than 200 times each. 
For task 1, we kept in the corpus all entries corresponding to the labels that had over 200 examples. This left us with 8 law area classes. Table ~\ref{tab:1} shows their distribution.

\begin{table}[h]
\begin{center}
\begin{tabular}{lr}
\hline 
\bf Law Area & \bf \# of cases\\ \hline
\iffalse
%distribution for 80,000 entry dataset
%CHAMBRE\_CIVILE 		& 75\\
CHAMBRE\_CIVILE\_1 		& 12,194 \\
CHAMBRE\_CIVILE\_2 		& 10,321 \\
CHAMBRE\_CIVILE\_3 		& 10,802  \\
CHAMBRE\_MIXTE 			& 170 \\
%CHAMBRES\_REUNI 		& 39  \\
CHAMBRE\_SOCIALE 		& 20,397 \\
CHAMBRE\_COMMERCIALE 	& 11,052 \\ 
CHAMBRE\_CRIMINNELLE	& 15,317\\
ASSEMBLEE\_PLENIERE 	& 391 \\
\fi
%distribution for 120,000 entry dataset
CHAMBRE\_SOCIALE 	& 33,139\\
CHAMBRE\_CIVILE\_1 	& 20,838 \\
CHAMBRE\_CIVILE\_2 	& 19,772\\
CHAMBRE\_CRIMINELLE & 18,476\\
CHAMBRE\_COMMERCIALE & 18,339\\
CHAMBRE\_CIVILE\_3 	& 15,095\\
ASSEMBLEE\_PLENIERE & 544\\
CHAMBRE\_MIXTE 		& 222\\
\hline
\end{tabular}
\end{center}
\caption{\label{tab:1} Distribution of Law Area labels over the Case Descriptions}
\end{table}

\noindent In establishing the ruling label set for predicting the case ruling (task 2), we were faced with a bigger challenge since, after the initial pre-processing, we were left with a list of 475 unique labels (from the initial 635). Looking at this list, we noticed that there were some entries which contained the same keyword repeated several times without having an overt interpretation for the repetition (e.g. \emph{cassation partielle rejet rejet cassation} appeared 145 times in the dataset) as opposed to other multi-word labels which could be easily interpreted (e.g \emph{cassation partielle sans renvoi} which appeared 1,015 times). 

\begin{figure*}[!ht]
\centering
\includegraphics[width=\textwidth]{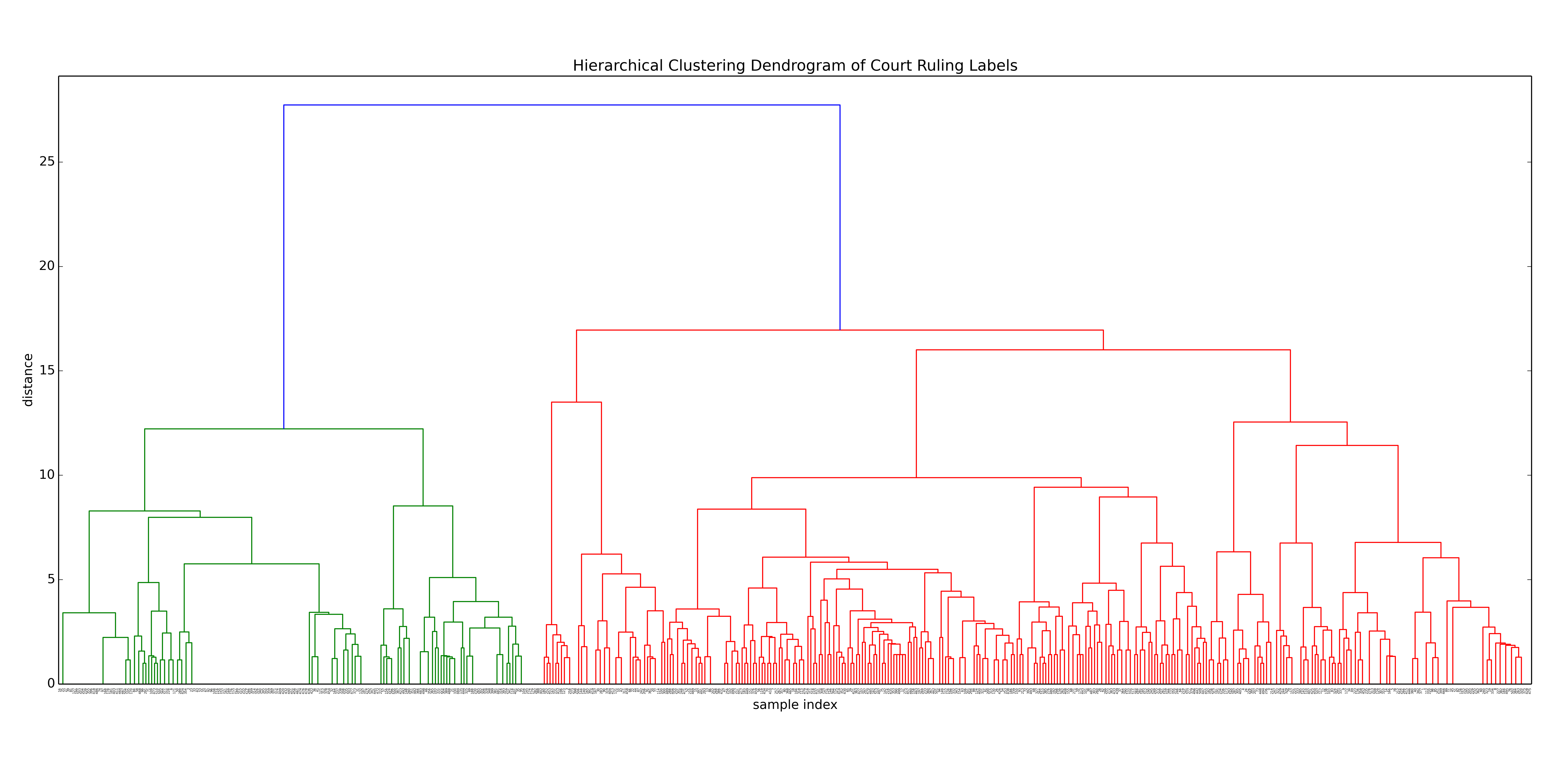}
\caption{Dendrogram showing hierarchical clustering of ruling labels}
\label{fig1}
\end{figure*}

An initial step, for better visualization of the ruling label space, was to do hierarchical clustering on the BOW occurrence vector representation for each label. We achieved this using Python's SciPy hierarchical functions with Ward distance (Figure \ref{fig1}).The results show good evidence for a high level clustering of labels into 6-8 groups. We then investigate what might be the basis of this clustering and determined that keeping only the  labels which had at least 200 examples was a good way to obtain this grouping.

On court ruling prediction, we carried out two sets of experiments. In the first one we considered only the first word within each label and only those labels which had over 200 entries in the corpus (first word setup). This lead to an initial set of 6 unique labels: \emph{cassation}, \emph{annulation}, \emph{irrecevabilite}, \emph{rejet}, \emph{non-lieu}, and \emph{qpc} (\emph{question prioritaire de constitutionnalité}). The motivation behind using the first word, rather than using a more complex approach for the identification of the "correct" label, was based on the fact that in French the adjective follows the noun and that the labels consisted only of nouns, adjectives, and stop words. %We coupled this knowledge  with the zipfian assumption that the most important information will always present itself at the beginning rather than the end and arrived at the decision of keeping the first word as label.

In the second set of experiments, we considered all labels which had over 200 dataset entries and this time we did not reduce them to their first word. %We call this the multi-word setup. 
Table ~\ref{tab:2} shows the distribution of the ruling labels with over 200 examples each. Italics were used here to emphasize those labels which do not have an overt semantic interpretation. An important observation here is that, in the full, multi-word label extraction setup, \emph{non-lieu} and \emph{qpc}, which are known to be valid decisions of the French Supreme Court, are not selected as final labels, unlike in the first-word setup. This happens because they appear at the beginning of several rare labels (e.g. non-lieu a statuer,  non-lieu a recevoir, qpc seule irrecevabilite, etc.). 
%rather than appearing on their own. 
Therefore, there are not enough instances in the dataset with these labels for these labels to be selected. A similar phenomenon occurs with the rest of the labels when comparing the first-word to the multi-word setup.

\begin{table}[h]
\begin{center}
\begin{tabular}{lr}
\hline 
\bf First-word ruling & \bf \# of cases\\ \hline
\iffalse
%distribution for 80,000 entry dataset
cassation 			& 39,842\\
rejet 				& 38,209  \\
irrecevabilite		& 1,933 \\
annulation			& 305 \\
non-lieu			& 158  \\ \hline
\fi
%distribution for 200,000 entry dataset
rejet & 68,516 \\
cassation & 53,813 \\
irrecevabilite & 2,737 \\
qpc & 409 \\
annulation & 377 \\
non-lieu & 246 \\
\hline
\bf Full ruling & \bf \# of cases\\ \hline
\iffalse
%distribution for 80,000 entry dataset
rejet 										& 37,750 \\
cassation 									& 27,375 \\
cassation sans renvoi 						& 1,804 \\
cassation partielle 						& 6,771 \\
cassation partielle sans renvoi  			& 869 \\ 
\emph{cassation partielle cassation} 		& 960 \\
\emph{cassation partielle rejet cassation} 	& 798 \\
\emph{cassation partielle cassation rejet} 	& 179 \\
irrecevabilite 								& 1,615 \\
annulation 									& 156\\
\fi
%distribution for 200,000 entry dataset
cassation & 37,659 \\
cassation sans renvoi & 2,078 \\
cassation partielle & 9,543 \\
cassation partielle sans renvoi & 1,015 \\
\emph{cassation partielle cassation} & 1,162 \\
\emph{cassation partielle rejet cassation} & 906 \\
rejet & 67,981 \\
irrecevabilite & 2,376 \\
\hline
\end{tabular}
\end{center}
\caption{\label{tab:2} Distribution of Case Ruling labels over the Case Descriptions}
\end{table}

\noindent Finally, for the temporal text classification (task 3), we initially considered the decade of the ruling and the case description. The distribution is shown on Table ~\ref{tab:3}, with the 1970 being the most prolific in cases.

As discussed in \newcite{zampieri2016lrec} the definition of time spans for supervised temporal text classification is often arbitrary. Given that most cases were dated after 1960 and previous decades had only a few cases each, we divided the dataset into 7 classes by grouping all cases before 1960 under one label. Secondly, we considered fine-grained intervals by dividing the dataset into 14 classes merging classes before 1920 as follows: 1830-1840, 1850-1860, 1870-1880, 1890-1910. 

\begin{table}[h]
\begin{center}
\begin{tabular}{lrlr}
\hline 
\bf Period & \bf \# of CR & \bf Period & \bf \# of CR \\ \hline
1880s			& 1 	& 	1870s	& 8		\\	
1810s			& 2 	& 	1880s	&	10	\\
1820s			& 2  	& 	1890s	&	8	\\
1830s			& 1   	& 	1910s	&	2	\\
1840s			& 4   	& 	1920s	&	17	\\ 
1850s			& 9   	& 	1930s	&	29	\\
1860s			& 9   	& 	1940s	&	15	\\
1950s			& 84 	& 	1960s	& 4,797 \\
1970s			& 23,964 & 1980s	& 18,233 \\
1990s			& 16,693 	& 2000s	& 12,577 \\
2010s		& 4,541\\

\hline
\end{tabular}
\end{center}
\caption{\label{tab:3} Distribution of Ruling Date labels over the Case Descriptions}
\end{table}

%The reason for employing the second (14 class) setup and not the full 21 classes, as can be seen from Table ~\ref{tab:3}, lies on the fact that the distribution of the 21 classes does not allow for a 10 fold cross validation, since there are some classes from the ones that were merged that have less than 10 examples.

\subsection{Masking and Feature Selection}
\label{sec:masking}

To emulate a real-world scenario in which a system would operate on a "draft" case description which does not indicate the desired target features to be predicted, we had to eliminate the occurrence of each word of the labels to predict from the text of the corresponding case description. 

For task 1, law area prediction, we eliminated all words contained in the respective label. For task 2, predicting the ruling, we initially eliminated from the case description all occurrences of the ruling word itself. We run ANOVA testing on the feature set used for classification (bag of words) and looked at the top 20 features to make sure that none of them could be construed as being directly linked to the label we were attempting to predict, so that a complete masking of the ruling was done within the case description text. In doing so, we realized the label was present both in its nominal form (e.g. {\em cassation, irrecevabilite}) and in its verbal forms (e.g. {\em casse, casser}). We eliminated these forms too. 

We finally investigated whether this technique of picking the top k classification features was good for identifying facts in the case description, aspects by which one would expect a lawyer to predict the judge's ruling. We did this by looking at the highest-ranking 20 word bigrams and trigrams from the feature set. What we  found instead were nouns with their articles (e.g. {\em la cause, le pourvoi}), prepositions with verbs and nouns (e.g. {\em pour etre, sur interpretation}), for bi-grams, and infinitival constructions (e.g. {\em et pour etre, occasion de faire}), for trigrams. 

Finally, for task 3, estimating the data of the case, we eliminate all digits from the case description. This has the disadvantage of removing digits that may refer to cited laws thus making the task even more challenging.

\section{Computational Approach}

We approach the tasks using a text classification system based on the scikit-learn implementation \cite{skl} of the LIBLINEAR SVM classifier \cite{liblinear}. As features, we investigate word unigrams (bag of words) and bigrams (bag of bigrams) frequencies to capture the appropriate differences between case descriptions. We extract these features using scikit-learn's CountVectorizer. 

Since these features rendered lower performance in temporal classification than in the first two tasks, we also look at other features as proposed in \newcite{niculae2014temporal} to improve the performance. Specifically, we couple BOW with the type-token ratio of each case description computed in the following way: 

%\iffalse
\begin{displaymath}
word\_type\_token = \frac{\#unique\_words}{\#total\_words}
\end{displaymath}
%\fi

\noindent As the dataset is imbalanced, we employ stratified 10-fold cross-validation %initial 
for all experiments, since this validation method maintains the initial distribution over each fold. We compare our scores against a random baseline classifier implemented in scikit-learn as the DummyClassifier which takes into consideration the dataset's initial distribution. We report average precision, recall, and f1 scores over all labels. The C hyperparameter for the linear SVM was set to 0.1 in all experiments employing SVMs.

\section{Results}
\label{sec:results}

In this section we report the results obtained for the three tasks all under the masking regimes described in Section \ref{sec:masking}: (1) predicting the law area of a case, (2) predicting the ruling of a case based on a case description, and (3) estimating the date of a case.

\subsection{Law Area}
\label{sec:area}

In the first experiment, we apply the SVM classifier to predict the law area of a case. Table ~\ref{tab:res-la1} shows the results  of this classifier applied to 8 classes containing at least 200 instances each presented in Table \ref{tab:2}.

\begin{center}
\begin{table}[!ht]
\centering
\begin{tabular}{ccccc} 
\hline 
\bf Model 		& \bf P & \bf R 	& \bf F1 	& \bf Acc.\\\hline
SVM 			& 90.9\%		& 90.2\%	& 90.3\% 	& 90.2\%\\
baseline 			& 17.7\%		& 17.7\%	& 17.7\%  & 17.\% \\
%bigram 	& 33\%		& 58\%	& 42\%	& 58\%\\
%dummy 2-gram	& 38\%		& 38\%	& 38\%  & 38\% \\
\hline
\end{tabular}
\caption{Classification results for the law area prediction task using Linear SVM on 8 classes}
\label{tab:res-la1}
\end{table}
\end{center}

%\vspace{-4mm}

The results show that on average our system is able to predict the law area of a case and court ruling with high precision, recall, and f1 score, well above those of the random baseline. 
%When we investigate the performance on the minority classes, we see a drop in performance from the average, but the scores are maintained well above those of the random baseline.

\subsection{Court Ruling}
\label{sec:ruling}

In this section we present the results obtained in the second task, ruling prediction based on a case description. The results are presented in Table ~\ref{tab:res-rul}. We report the scores of the experiments when run on the first-word (6 classes) as well as multi-word setups (8 classes) for label extraction discussed in Section \ref{sec:label}. 

We observe an apparent 6 percentage points decrease in average scores when the classifier is trained on the dataset with more classes. This is in tune with the characteristic of classifiers such as SVM which suffer from imbalanced data and is to a certain extent expected since the class imbalance is significant. However, it is important to note that the drop is only apparent, since the increase in number of classes leads to a decrease in the random baseline performance and thus the difference between the baseline scores and our method actually grows by 4 percentage points from the first-word setup. 

\begin{table}[!ht]
\centering
\scalebox{0.94}{
\begin{tabular}{lllll} 
\hline 
 \bf Model & \bf P & \bf R & \bf F1 	& \bf Acc.\\\hline
6 cls SVM		& 97.1\%	& 96.9\%	& 97.0\% 	& 96.9\% \\
6 cls baseline	& 47.7\%	& 47.7\%	& 47.7\%  	& 47.7\% \\\hline
8 cls SVM		& 93.2\%	& 92.8\%	& 92.7\% 	& 92.8\% \\
8 cls baseline 	& 40.6\%	& 40.6\%	& 40.6\%	& 40.6\% \\
\hline
\end{tabular}
}
\caption{Classification results for the ruling prediction task using Linear SVM}
\label{tab:res-rul}
\end{table}

\noindent In terms of previous work, unfortunately a systematic and thorough comparison with \newcite{Katz14} and \newcite{wongchaisuwat2016} is not possible since we are not using the same corpus nor working on the same language as these two papers. Even so, our method appears to surpass both, in terms of f1 score, in predicting the ruling of a court, based on previous examples. One main difference might be the judicial system which is known to be more predictable (offering the judges less interpretation freedom) in the case of the French Supreme Court.

\subsection{Temporal Text Classification}
\label{sec:temporal}

\begin{table*}
\centering
\begin{tabular}{llllll} 
\hline 
\bf Subtask & \bf Model & \bf Precision & \bf Recall & \bf F1 	& \bf Accuracy\\\hline
7-class & SVM 1-gram	& 69.9\%	& 68.3\%	& 68.2\% 	& 68.3\% \\
7-class & SVM 2-gram	& \bf75.9\%	& \bf74.3\%	& \bf73.2\% & \bf74.3\% \\
7-class & baseline 		& 19.2\%	& 19.2\%	& 19.2\%  	& 19.2\% \\ \hline
14-class & SVM 1-gram 	& 69.1\%	& 68.6\%	& 68.5\% 	& 68.6\% \\
14-class & SVM 2-gram 	& \bf75.6\%	& \bf74.2\%	& \bf73.9\% & \bf74.2\% \\
14-class & baseline 	& 19.1\%	& 19.1\%	& 19.1\%	& 19.1\% \\ \hline
\end{tabular}
\caption{Classification results for temporal prediction using Linear SVM}
\label{tab:res-date}
\end{table*}

For the third task, estimating the date of case and ruling, we use the same approach as previous experiments, a linear SVM classifier trained on bag of unigrams and bag of bigrams as features. Results in two settings, one containing 7 classes and the other containing 14 classes, are reported in Table \ref{tab:res-date} 
%Here, again, we notice a drop in performance of our classifier when comparing its scores to those obtained by the same classifier (with same feature extraction method, but different feature space) on the ruling prediction task. However, this apparent drop is no longer evident when comparing our classifier scores with those of the dummy classifier (the stratified baseline) on the same task and same label extraction setup. In fact, the difference between the random baseline and our method for the third task, in both class settings, is the same as that obtained for the second task, in the 8-class setup.

The general tendency of traditional supervised classification algorithms is to increase their performance as the number of classes or imbalance between classes decreases. Our experiments show that we manage to preserve the difference between the baseline performance and that of our system on different tasks (ruling prediction and temporal classification), with varying number of classes and initial distributions, which suggests that these techniques are robust for our purpose. However, from a user perspective, where error rate needs to be low, we expect this observation to not be useful and we therefore also run the SVM experiments with type-token ratios as features. On their own, they were able to reach a little above the random baseline (43\% f1 score vs. 19\% for the random). Interestingly, type-token ratio did not increase the performance of the classifier when combined with BOW.

\section{Conclusions and Future Work}

In this paper we investigated the application of text classification methods to legal texts from the French Supreme Court. To the best of our knowledge, this is the first work to: (1) apply text classification to predict the rulings on a French dataset, (2) carry out temporal text classification experiments on legal texts. The paper also reports high performance in the task of predicting court rulings.

We showed that a linear SVM classifier trained on BOW can obtain high f1 scores in predicting the law area and the ruling of a case, given the case description. Estimating the date of cases turned out to be more difficult to learn using bag of words and lexical richness features (type-token ratio), but this may be due to the highly imbalanced dataset (i.e. too few examples from the minority classes) or to the possible fact that the language used by judges of the French Supreme Court over the years has not changed much. This final observation is worth further investigation.

We also looked at ways of masking the case description to convey as little information as possible regarding the ruling itself making the task more challenging. This method showed that the word bigrams and trigrams deemed to be the most salient in predicting the ruling are not actually tied to any factual information particular to one case, but more related to formulaic expressions typical for a particular ruling. In future work, we would like to extend this investigation to the sentence level and see if the sentences that are considered most effective in predicting the ruling are of factual nature. 

Our work is proof of concept that text classification techniques can be used to provide valuable assistive technology for law professionals in obtaining guidance and orientation from large corpora of previous court rulings. In the future we would like to investigate more sophisticated methods of masking features in the original text data that explicitly list and ``give away'' the desired target prediction to simulate realistic application scenarios, where text classification predicts the target features from ``draft''  case descriptions that do not yet contain the target predictions.

Finally, we would like to improve the performance of our system by exploring the combination of other features and the use of ensembles and meta-classifiers which proved to achieve high performance in other text classification tasks \cite{malmasi2016predicting}.

\section*{Acknowledgements}

This work was carried out while the first and the second author, Octavia-Maria \c{S}ulea and Marcos Zampieri, were at the German Research Center for Artificial Intelligence (DFKI).

%\bibliography{acl2017}
\bibliography{acl2017}

\begin{thebibliography}{}
\expandafter\ifx\csname natexlab\endcsname\relax\def\natexlab#1{#1}\fi

\bibitem[{Aletras et~al.(2016)Aletras, Tsarapatsanis, Preo{\c{t}}iuc-Pietro,
  and Lampos}]{aletras2016predicting}
Nikolaos Aletras, Dimitrios Tsarapatsanis, Daniel Preo{\c{t}}iuc-Pietro, and
  Vasileios Lampos. 2016.
\newblock {Predicting Judicial Decisions of the European Court of Human Rights:
  A Natural Language Processing Perspective}.
\newblock {\em PeerJ Computer Science\/} 2:e93.

\bibitem[{Balahur et~al.(2014)Balahur, Mihalcea, and Montoyo}]{balahur14}
Alexandra Balahur, Rada Mihalcea, and Andr{\'{e}}s Montoyo. 2014.
\newblock {Computational Approaches to Subjectivity and Sentiment Analysis:
  Present and Envisaged Methods and Applications}.
\newblock {\em Computer Speech {\&} Language\/} 28(1):1--6.

\bibitem[{Barr{\'o}n-Cede{\~n}o et~al.(2013)Barr{\'o}n-Cede{\~n}o, Vila,
  Mart{\'\i}, and Rosso}]{barron2013plagiarism}
Alberto Barr{\'o}n-Cede{\~n}o, Marta Vila, M~Ant{\`o}nia Mart{\'\i}, and Paolo
  Rosso. 2013.
\newblock {Plagiarism Meets Paraphrasing: Insights for the Next Generation in
  Automatic Plagiarism Detection}.
\newblock {\em Computational Linguistics\/} 39(4):917--947.

\bibitem[{Boella et~al.(2011)Boella, Caro, , and Humphreys}]{Boella:2011}
Guido Boella, Luigi~Di Caro, , and Llio Humphreys. 2011.
\newblock {Using Classification to Support Legal Knowledge Engineers in the
  Eunomos Legal Document Management System}.
\newblock In {\em Proceedings of JURISIN\/}.

\bibitem[{Ciobanu et~al.(2017)Ciobanu, Zampieri, Malmasi, and
  Dinu}]{ciobanu2017including}
Alina~Maria Ciobanu, Marcos Zampieri, Shervin Malmasi, and Liviu~P Dinu. 2017.
\newblock {Including Dialects and Language Varieties in Author Profiling}.
\newblock In {\em Proceedings of CLEF\/}.

\bibitem[{De~Vel et~al.(2001)De~Vel, Anderson, Corney, and
  Mohay}]{de2001mining}
Olivier De~Vel, Alison Anderson, Malcolm Corney, and George Mohay. 2001.
\newblock {Mining E-mail Content for Author Identification Forensics}.
\newblock {\em ACM Sigmod Record\/} 30(4):55--64.

\bibitem[{Dinu et~al.(2012)Dinu, Niculae, and {\c{S}}ulea}]{dinu2012pastiche}
Liviu~P Dinu, Vlad Niculae, and Octavia-Maria {\c{S}}ulea. 2012.
\newblock {Pastiche Detection Based on Stopword Rankings: Exposing
  Impersonators of a Romanian Writer}.
\newblock In {\em Proceedings of the Workshop on Computational Approaches to
  Deception Detection\/}.

\bibitem[{Fan et~al.(2008)Fan, Chang, Hsieh, Wang, and Lin}]{liblinear}
Rong-En Fan, Kai-Wei Chang, Cho-Jui Hsieh, Xiang-Rui Wang, and Chih-Jen Lin.
  2008.
\newblock {LIBLINEAR: A Library for Large Linear Classification}.
\newblock {\em Journal of Machine Learning Research\/} 9:1871--1874.

\bibitem[{Farzindar and Lapalme(2004)}]{farzindar2004legal}
Atefeh Farzindar and Guy Lapalme. 2004.
\newblock {Legal Text Summarization by Exploration of the Thematic Structures
  and Argumentative Roles}.
\newblock {\em Proceedings of the Text Summarization Branches Out Workshop\/} .

\bibitem[{Galgani et~al.(2012)Galgani, Compton, and Hoffmann}]{galgani12}
Filippo Galgani, Paul Compton, and Achim Hoffmann. 2012.
\newblock {Combining Different Summarization Techniques for Legal Text}.
\newblock In {\em Proceedings of the Hybrid Workshop\/}.

\bibitem[{Ganascia et~al.(2014)Ganascia, Glaudes, and Lungo}]{ganascia14}
Jean-Gabriel Ganascia, Pierre Glaudes, and Andrea~Del Lungo. 2014.
\newblock Automatic detection of reuses and citations in literary texts.
\newblock {\em Digital Scholarship in the Humanities\/} 29(3).

\bibitem[{Gebre et~al.(2013)Gebre, Zampieri, Wittenburg, and
  Heskes}]{gebre2013}
Binyam~Gebrekidan Gebre, Marcos Zampieri, Peter Wittenburg, and Tom Heskes.
  2013.
\newblock {Improving Native Language Identification with TF-IDF Weighting}.
\newblock In {\em Proceedings of the BEA Workshop\/}.

\bibitem[{Gon{\c{c}}alves and Quaresma(2005)}]{gonccalves2005evaluating}
Teresa Gon{\c{c}}alves and Paulo Quaresma. 2005.
\newblock {Evaluating Preprocessing Techniques in a Text Classification
  Problem}.
\newblock In {\em Proceedings of the Conference of the Brazilian Computer
  Society\/}.

\bibitem[{Hachey and Grover(2006)}]{hachey2006extractive}
Ben Hachey and Claire Grover. 2006.
\newblock {Extractive Summarisation of Legal Texts}.
\newblock {\em Artificial Intelligence and Law\/} 14(4):305--345.

\bibitem[{Katz et~al.(2014)Katz, II, and Blackman}]{Katz14}
Daniel~Martin Katz, Michael J.~Bommarito II, and Josh Blackman. 2014.
\newblock {Predicting the Behavior of the Supreme Court of the United States: A
  General Approach}.
\newblock {\em CoRR\/} abs/1407.6333.

\bibitem[{Malmasi et~al.(2016)Malmasi, Zampieri, and
  Dras}]{malmasi2016predicting}
Shervin Malmasi, Marcos Zampieri, and Mark Dras. 2016.
\newblock {Predicting Post Severity in Mental Health Forums}.
\newblock In {\em Proceedings of the CLPsych Workshop\/}.

\bibitem[{Nguyen et~al.(2013)Nguyen, Gravel, Trieschnigg, and
  Meder}]{nguyen2013}
Dong-Phuong Nguyen, Rilana Gravel, RB~Trieschnigg, and Theo Meder. 2013.
\newblock {``How old do you think I am?'' A Study of Language and Age in
  Twitter}.
\newblock In {\em Proceedings of ICWSM\/}.

\bibitem[{Niculae et~al.(2014)Niculae, Zampieri, Dinu, and
  Ciobanu}]{niculae2014temporal}
Vlad Niculae, Marcos Zampieri, Liviu~P Dinu, and Alina~Maria Ciobanu. 2014.
\newblock {Temporal Text Ranking and Automatic Dating of Texts}.
\newblock {\em Proceedings of EACL\/} .

\bibitem[{Palau and Moens(2009)}]{Palau:2009}
Raquel~Mochales Palau and Marie-Francine Moens. 2009.
\newblock {Argumentation Mining: The Detection, Classification and Structure of
  Arguments in Text}.
\newblock In {\em Proceedings of the ICAIL\/}.

\bibitem[{Pedregosa et~al.(2011)Pedregosa, Varoquaux, Gramfort, Michel,
  Thirion, Grisel, Blondel, Prettenhofer, Weiss, Dubourg, Vanderplas, Passos,
  Cournapeau, Brucher, Perrot, and Duchesnay}]{skl}
F.~Pedregosa, G.~Varoquaux, A.~Gramfort, V.~Michel, B.~Thirion, O.~Grisel,
  M.~Blondel, P.~Prettenhofer, R.~Weiss, V.~Dubourg, J.~Vanderplas, A.~Passos,
  D.~Cournapeau, M.~Brucher, M.~Perrot, and E.~Duchesnay. 2011.
\newblock Scikit-learn: Machine learning in {P}ython.
\newblock {\em Journal of Machine Learning Research\/} 12:2825--2830.

\bibitem[{P{\'{e}}rez{-}Rosas and Mihalcea(2015)}]{rada15}
Ver{\'{o}}nica P{\'{e}}rez{-}Rosas and Rada Mihalcea. 2015.
\newblock {Experiments in Open Domain Deception Detection}.
\newblock In {\em Proceedings of EMNLP\/}.

\bibitem[{Sulea and Dichiu(2015)}]{sulea2}
Octavia{-}Maria Sulea and Daniel Dichiu. 2015.
\newblock {Automatic Profiling of Twitter Users Based on Their Tweets: Notebook
  for PAN at CLEF 2015}.
\newblock In {\em Proceedings of CLEF\/}.

\bibitem[{Sumner et~al.(2012)Sumner, Byers, Boochever, and Park}]{dark-triad}
Chris Sumner, Alison Byers, Rachel Boochever, and Gregory~J. Park. 2012.
\newblock {Predicting Dark Triad Personality Traits from Twitter Usage and a
  Linguistic Analysis of Tweets}.
\newblock In {\em Proceedings of ICMLA\/}.

\bibitem[{Wongchaisuwat et~al.(2016)Wongchaisuwat, Klabjan, and
  McGinnis}]{wongchaisuwat2016}
Papis Wongchaisuwat, Diego Klabjan, and John~O McGinnis. 2016.
\newblock {Predicting Litigation Likelihood and Time to Litigation for
  Patents}.
\newblock {\em arXiv preprint arXiv:1603.07394\/} .

\bibitem[{Zampieri et~al.(2016)Zampieri, Malmasi, and Dras}]{zampieri2016lrec}
Marcos Zampieri, Shervin Malmasi, and Mark Dras. 2016.
\newblock {Modeling Language Change in Historical Corpora: The Case of
  Portuguese}.
\newblock In {\em Proceedings of LREC\/}.

\end{thebibliography}
\bibliographystyle{acl_natbib}

\end{document}